\documentclass[sigconf]{acmart}
\usepackage{algpseudocode}
\usepackage{booktabs}
\usepackage{multirow}
\usepackage{multicol}
\usepackage{xspace}
\AtBeginDocument{%
  \providecommand\BibTeX{{%
    \normalfont B\kern-0.5em{\scshape i\kern-0.25em b}\kern-0.8em\TeX}}}

\usepackage{bm}
\usepackage{multirow}
\usepackage{mathtools}
\usepackage{subcaption}
\usepackage{paralist}
\usepackage[linesnumbered, ruled, vlined]{algorithm2e}
\SetKwInOut{Input}{Input}
\SetKwInOut{Output}{Output}

\newtheorem{remark}{\noindent \textbf{Remark}}
\newtheorem{proof_}{\noindent \textbf{Hypothesis}}

\usepackage{color}

\newcommand{\bizhen}[1]{{\color{black}#1}}

\setcopyright{acmcopyright}
\copyrightyear{2022}
\acmYear{2022}
\acmDOI{XXXXXXX.XXXXXXX}

\acmConference[]{Work in progress}
\acmPrice{}
\acmISBN{978-1-4503-XXXX-X/18/06}

\begin{document}
\acmBooktitle{}

\title{Relphormer: Relational Graph Transformer for\\ Knowledge Graph Representations}

\author{
Zhen Bi$^{12*}$, 
Siyuan Cheng$^{12*}$, 
Jing Chen$^{12*}$, 
Xiaozhuan Liang$^{12}$, 
Feiyu Xiong$^{3}$, 
Ningyu Zhang$^{12\dagger}$
}

\affiliation{
$^1$Zhejiang University \country{China} \\
$^2$Zhejiang University - Ant Group Joint Laboratory of Knowledge Graph\country{China} \\
$^3$Alibaba Group\country{China}
}

\email{
{bizhen\_zju,zhangningyu}@zju.edu.cn
}

\begin{abstract}
Transformers have achieved remarkable performance in widespread fields, including natural language processing, computer vision and graph mining. However, vanilla Transformer architectures have not yielded promising improvements in the Knowledge Graph (KG) representations, where the translational distance paradigm dominates this area. Note that vanilla Transformer architectures struggle to capture the intrinsically heterogeneous structural and semantic information of knowledge graphs. To this end, we propose a new variant of Transformer for knowledge graph representations dubbed Relphormer. Specifically, we introduce Triple2Seq which can dynamically sample contextualized sub-graph sequences as the input to alleviate the heterogeneity issue. We propose a novel structure-enhanced self-attention mechanism to encode the relational information and keep the semantic information within entities and relations. Moreover, we utilize masked knowledge modeling for general knowledge graph representation learning, which can be applied to various KG-based tasks including knowledge graph completion, question answering, and recommendation. Experimental results on six datasets show that Relphormer can obtain better performance compared with baselines\footnote{Code is available in \url{https://github.com/zjunlp/Relphormer}.}.

\end{abstract}



\maketitle

\section{Introduction}

\begin{figure}[!t]
\centering

\includegraphics[width=0.45\textwidth]{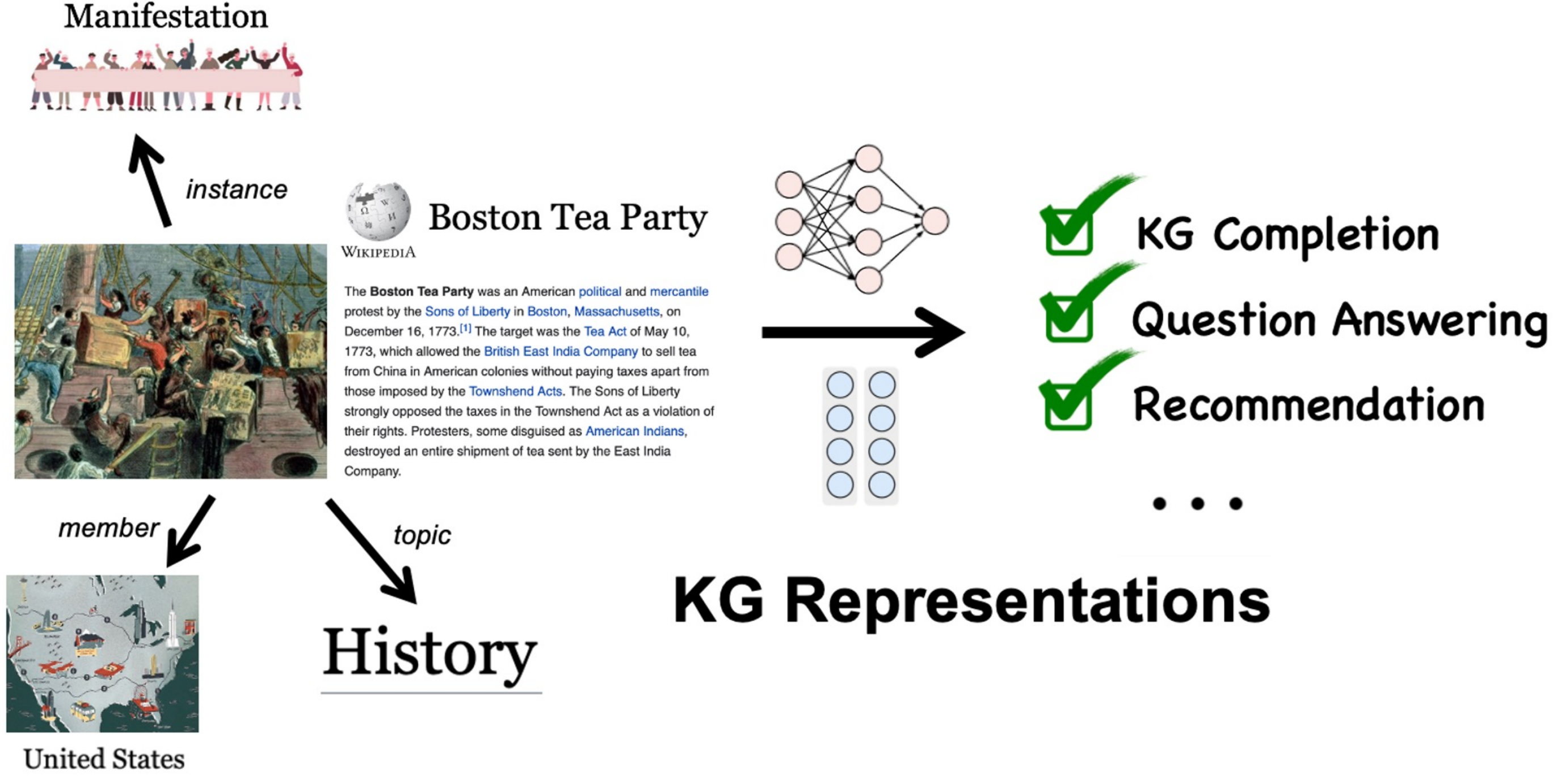}
\caption{
Examples from WN18RR and Wikipedia.
KG representations can be applied to various tasks.}
\label{KG representation}
\end{figure}

In recent years, Knowledge Graphs (KGs) have shown effectiveness in information retrieval and are widely applied to a variety of applications such as search engine \cite{xiong2017explicit}, recommendation \cite{liu2021reinforced}, time series prediction \cite{DBLP:conf/www/DengZZCPC19}, natural language understanding \cite{zhang2019ernie,yao2023editing,DBLP:journals/corr/abs-2210-00105}.
As shown in Figure \ref{KG representation}, KG representation learning aims to project the entities and relations into a continuous low-dimensional vector space, which can implicitly promote computations of the reasoning between entities and has been proved helpful for knowledge-intensive tasks \cite{DBLP:journals/corr/abs-2210-00305,DBLP:journals/tnn/JiPCMY22,DBLP:journals/kbs/WuKYQW22,DBLP:journals/tkde/GuoZQZXXH22}.
Previous KG representation learning methods,  such as TransE \cite{TransE}, ComplEx \cite{ComplEx} and RotatE \cite{RotatE},  embed the relational knowledge into a vector space and then optimize the target object by leveraging a pre-defined scoring function to those vectors. 
However, it is rather challenging to encode all information (e.g., semantic and structural) about an entity into a single vector. 
To this end, several works utilize graph neural networks (GNN) \cite{DBLP:conf/esws/SchlichtkrullKB18,DBLP:conf/www/Zhang0YW22} or attention-based approaches \cite{DBLP:conf/kdd/Wang00LC19} to learning representations based on both entities and their graph context.
However, these methods are still restricted in expressiveness regarding the shallow network architectures. 

The Transformer \cite{transformers}, acknowledged as the most powerful neural network in modeling sequential data, has achieved success in representing natural language \cite{BERT} and computer vision \cite{DBLP:conf/iclr/DosovitskiyB0WZ21}. 
Besides, there are many attempts to leverage the Transformer into the graph domain. 
The majority of these approaches \cite{Graphormer,Gophormer} apply Transformer on the entire graph and enhance the vanilla feature-based attention mechanism with topology-enhanced attention mechanism and structural encoding or leveraging Transformer through ego-graphs with proximity-enhanced attention. 
Note that different from the pure graph, KGs are heterogeneous graphs consisting of multiple types of nodes.
Therefore, \textbf{it is still an open question whether Transformer architecture is suitable to model KG representations and how to make it work for various tasks}.
Concretely, there are three nontrivial challenges for fully applying Transformer architecture to KGs as follows:
\begin{itemize}
\item  \textbf{Heterogeneity for edges and nodes.}
Note that  KGs are relational graphs with semantically enriched edges. 
Multiple edges have different relational information, resulting in distinct heterogeneous representations. 
Previous Transformer architecture usually applies implicit structure inductive bias (typically single-hop neighbors) as constraints, ignoring the essential graph structures of KG.

\item  \textbf{Topological structure and textual description.}
KGs are text-rich networks with two types of knowledge for each node, that is, topological structure and textual descriptions.
Different nodes have unique topological and textual features.
For example, given an entity-relation pair $(h,r)$, characterizing the structural context and textual descriptions will provide valuable information when inferring the entity node from $h$ and $t$.
However, previous Transformer architecture usually treats all entities and relations as plain tokens, missing the essential structural information.

\item  \textbf{Task Optimization Universality.} 
Note that most previous studies \cite{TransE} follow the paradigm with a pre-defined scoring function for knowledge embeddings. 
However, such a strategy requires optimizing different objects for entity/relation prediction and costly scoring of all possible triples in inference. 
Meanwhile, previous KG representation learning methods struggle to provide unified learned representations for various KG-based tasks, such as knowledge graph completion, question answering and recommendation.
Therefore, it is intuitive to design a new technical solution for knowledge graph representation.

\end{itemize}

In this study, to address the above-mentioned issues, we explore the Transformer architecture for knowledge graph representation and propose \textbf{Rel}ational Gra\textbf{ph} Transf\textbf{ormer} (\textbf{Relphormer}).

\textbf{First}, we propose Triple2Seq  to address the heterogeneity for edges and nodes, which sample contextual sub-graphs as input sequences for the model.
Specifically, we regard relations as normal nodes in the sub-graphs and then feed the contextualized sub-graphs into the transformer module.
We use a dynamic sampling strategy to maximally preserve the localized contextual information and semantics.
\textbf{Second}, we propose a structure-enhanced mechanism to handle the topological structure and textual description.
With our designed module, the model can simultaneously encode the textual features while preserving the structural information.
Note that we not only focus on knowledge graph representation, such a mechanism is readily applicable to other kinds of Transformer-based approaches to incorporate any structural bias.
\textbf{Finally}, we propose a masked knowledge modeling mechanism for unified knowledge graph representation learning.
Inspired by Masked Language Modeling in natural language processing, we introduce a unified optimization object of predicting masked entities as well as relation tokens in the input sequence. 
In such a way, we can simply leverage a unified optimization object for entity and relation prediction in knowledge graph completion.
Moreover, we can utilize learned KG representations for various KG-based tasks including question answering and recommendation tasks.

We conduct experiments to evaluate our Relphormer in various tasks.
Specifically, for KG completion, we use FB15K-237 \cite{FB15K237}, WN18RR \cite{WN18RR} and UMLS \cite{UMLS} for entity prediction and choose FB15K-237 and WN18RR for relation prediction.
We also evaluate Relphormer in two KG-based tasks: FreebaseQA \cite{FreebaseQA} and WebQuestionSP \cite{WebQuestionsSP} for Question Answering (QA) and ML-20m \cite{MovieLens} for the recommendation task.
Experimental results illustrate that the proposed Relphormer can yield better performance and require less inference time.

\begin{itemize}
    \item We propose Relphormer, which is the Transformer architecture variant for knowledge graph representation. 
    Our work may open up new avenues for representing heterogeneous graphs through Transformer. 
    
    \item We propose Triple2Seq  to address the heterogeneity for edges and nodes and a structure-enhanced mechanism to handle the topological structure and textual description.
    
    \item Experimental results on six benchmark datasets demonstrate that our Relphormer can achieve better performance compared with baselines and is particularly beneficial in handling complex structure knowledge.
\end{itemize}

\begin{figure*}[!t]
\centering

\includegraphics[width=0.9\textwidth]{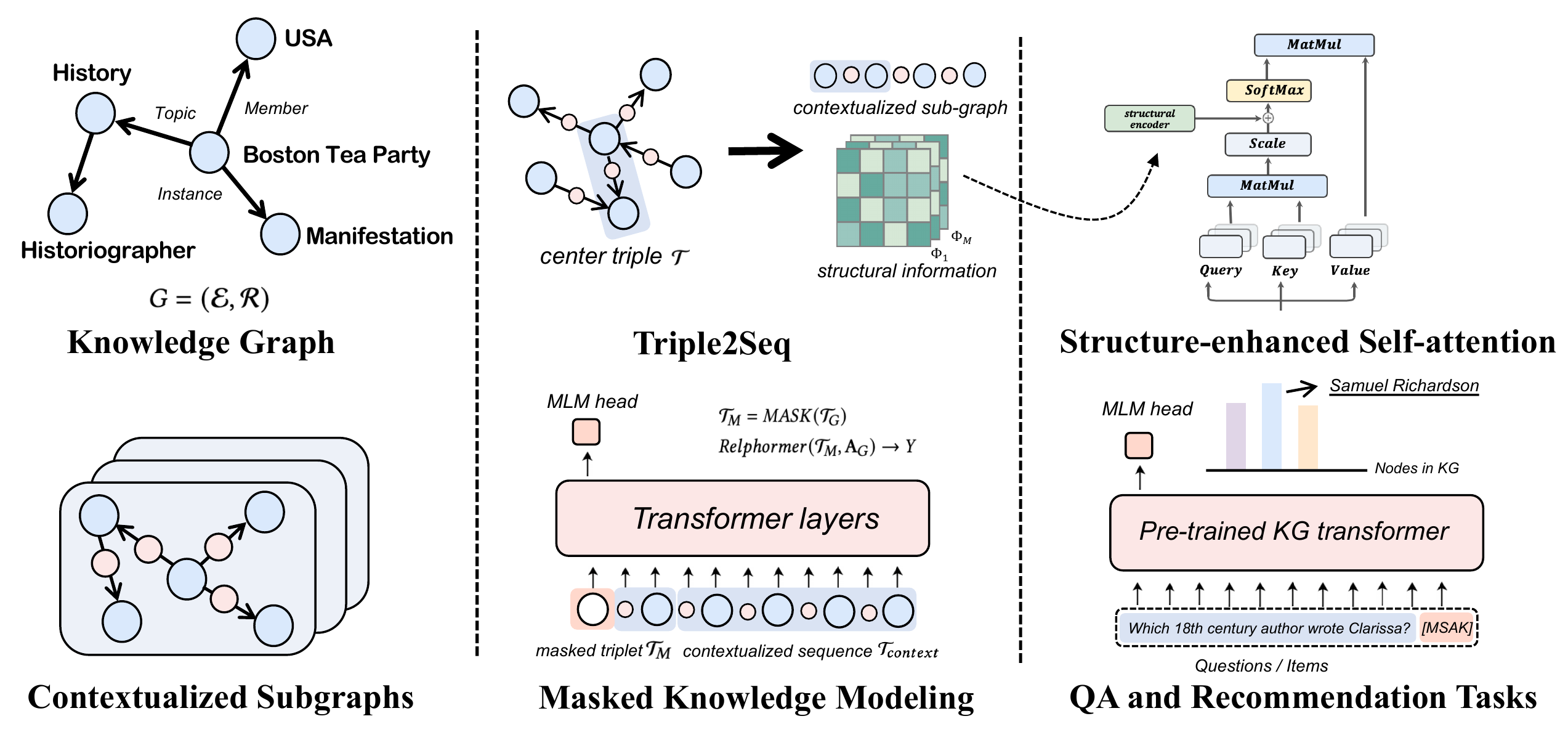}
\caption{
The model architecture of Relphormer. 
The contextualized sub-graph is sampled with Triple2Seq (Section \ref{sec:Triple2Seq}), and then it will be converted into textual sequences while maintaining its sub-graph structure.
For the transformer architecture, we design a novel structure-enhanced mechanism (Section \ref{sec:Structural_bias}) to preserve the structure feature.
Finally, we utilize masked knowledge modeling to pre-train Relphormer (Section \ref{sec:MKM}) and apply to KG-based tasks (Section \ref{sec:downtream_task}). 
}
\label{arc}
\end{figure*}

\section{Related work}
\subsection{Knowledge Graph Representation Learning}
To date, most existing KG completion methods such as TransE \cite{TransE}, TransD \cite{TransD}, TransR \cite{TransR} and TransH \cite{TransH}, utilize the embedding-based paradigm.
DistMult \cite{DistMult} uses a simple bilinear formulation and introduces a novel approach that utilizes the learned relation embeddings to mine logical rules.
ComplEx \cite{ComplEx} designs the composition of complex embeddings that can handle a large variety of binary relations.
DenseE \cite{DenseE} also provides an improved modeling scheme for knowledge graph representations by using the complex composition patterns of relations.
RotatE \cite{RotatE} is a practical approach to model and infer various relation patterns for knowledge graph embedding.
Apart from traditional translational distance methods, there has been considerable interest in using graph convolution networks \cite{GCN, GAT}. 
RGCN \cite{RGCN} deals with the highly multi-relational data characteristic of realistic KGs.
CompGCN \cite{CompGCN} proposes a novel graph convolutional framework that jointly embeds both nodes and relations.
\cite{DBLP:conf/www/Zhang0YW22} further explores the real effect of GCNs in KGC and finds that the transformations for entity representations are responsible for the performance improvements.

KG representations can boost the performance of KG-based tasks.
\cite{HittER} leverages their pre-trained HittER model and proves its performance on two QA datasets.
\cite{kgTransformer} formulates complex logical query answering as a masked prediction problem and validates it on the downstream reasoning datasets.
KGT5 \cite{KGT5} is a sequence-to-sequence method for both KGC and QA tasks.
However, these methods are still restricted in expressiveness regarding the model architectures.

\subsection{Transformer for Graphs}
Recently, a great variety of Transformers has been proposed to encode graph-structured data.
GraphTrans \cite{GraphTrans} uses the Transformer-based self-attention to learn long-range pairwise relations, where a novel readout mechanism is designed to obtain a global graph embedding.
Grover \cite{Grover} carefully designs self-supervised tasks in node-, edge- and graph-level, which can learn rich structural and semantic information of molecules from enormous unlabelled data.
It can address the issues of insufficient labeled molecules samples and poor generalization capability on large-scale molecular data.
Graphormer \cite{Graphormer} proposes several simple yet effective structural encoding methods to better model graph-structured data and mathematically proves that many popular GNN variants could be covered as the special cases of Graphormer. 

To overcome the limitation due to the independent modeling of textual features, GraphFormers \cite{GraphFormers} designs a new architecture where layerwise GNN components are nested alongside the transformer blocks of language models.
Gophormer \cite{Gophormer} applies transformers on ego-graphs instead of full graphs to alleviate severe scalability issues on the node classification task.
Heterformer \cite{Heterformer} proposes a heterogeneous GNN-nested transformer that blends GNNs and PLMs into a unified model.
GraphiT \cite{GraphiT} views graphs as sets of node features and incorporates structural and positional information into a transformer architecture, which outperforms representations learned with GNNs.

\subsection{Transformer for Knowledge Graph}
Various efforts have been devoted to exploring Transformer in the knowledge graph representation learning. 
KG-BERT \cite{KG-BERT} treats triples in knowledge graphs as textual sequences and uses pre-trained language models for knowledge graph completion.
StAR \cite{StAR} designs a structure-augmented text representation learning framework for efficient knowledge graph completion.
PKGC \cite{PKGC} proposes to convert each triple and its information into prompt sentences, which are further fed into PLMs for classification tasks.
HittER \cite{HittER} uses entity and contextualization and designs a hierarchical Transformer for knowledge graph embeddings.
kgTransformer \cite{kgTransformer} presents the Knowledge Graph Transformer with masked pre-training and fine-tuning strategies.
Besides, there are also some recently proposed promising works such as LP-BERT \cite{LP-BERT}, SimKGC \cite{DBLP:conf/acl/0046ZWL22}.

Unlike those approaches, Relphormer focuses on addressing key issues of Transformer for KGs and paves a new step for architecture designing. 
We propose a novel framework for unified knowledge representation learning and utilize the pre-trained knowledge representations for KG-based tasks.

\section{Proposed Approach: Relphormer}
\subsection{Preliminaries}
In this paper, we target the task for KG representation learning, which includes knowledge graph completion and KG-enhanced tasks of question answering and recommendation.
We detail the general notations in Table \ref{table:Definition_Notation} and illustrate the  model architecture of Relphormer in Figure \ref{arc}.

\begin{table}[!t]
    \caption{Notation used in this paper.}
        
    \centering
        \begin{tabular}{cc}
            \toprule
            \textbf{Symbol}  & \textbf{Description}\\
            \midrule
            $G$ & Knowledge Graph
            \\    
            $\mathcal{E}$  & Entity set
            \\  
            $\mathcal{R}$  & Relation set
            \\  
            $V$  & Nodes in the Knowledge Graph
            \\  
            $\mathcal{T}$  & triple $(v_s, v_p, v_o)$
            \\  
            $\mathcal{T}_G$ & contextualized sub-graph of the triple $\mathcal{T}$
            \\
            $\mathcal{T}_{M}$ & masked contextualized sub-graph $\mathcal{T}_G$
            \\ 
            $\mathbf{A}_G$  & Adjacency matrix of $\mathcal{T}_G$
            \\
            $M$  & The order of the contextualized sub-graph
            \\
            $\mathcal{T}_{context}$ & neighborhood nodes surrounding $\mathcal{T}$
            \\
            \bottomrule
        \end{tabular}
    \label{table:Definition_Notation}
\end{table}%
 
Let $\mathcal{G} = (\mathcal{E}, \mathcal{R})$ denote a KG with a entity set $\mathcal{E}$ and relation set $\mathcal{R}$.
We denote $V$ as the node set which consists of \textbf{both entities and relations}, that is $V = \mathcal{E} \cup \mathcal{R}$.
We define an adjacency matrix $\mathbf{A} = \{ a_{uv}\} \in \{ 0, 1 \}^{\left | V  \right |  \times \left | V \right | } $ where $a_{uv}=1$ if node pair $(u,v)$ is connected and $a_{uv}=0$ otherwise. 
It should be noted that the node pair $(u,v)$ can be interactions among entities and relations.
We define the triple as  $(v_{{s}}, v_{p}, v_{o} )$ or $\mathcal{T}$ in short, which acts as the "center triple" in the sampled sub-graphs.
$\mathcal{T}_G$ is the contextualized sub-graph surrounding the center triple (including the center triple itself).
KG representation learning is to learn a mapping $f:\mathcal{T}_G \to Y$, where  $\mathbf{Y} \in \mathbb{R}^{ \left |  \mathcal{E}  \right | + \left |  \mathcal{R}  \right | }$ are the label sets.

For the QA task, each question is labeled with an answering entity.
Given the question, the goal of the QA task is to find the correct answer.
Since some candidate entities are in the KG, the potential associations of entities will help improve the performance of the QA task.
For the recommendation task, given a set of items $(i_1, i_2, ..)$ for each user $u_i$, the goal is to predict the potential items in which the user might be interested.
The KG here refers to the item-centric KG, which organizes external item attributes with different types of entities and relations.
We hold the view that knowledge representation should be unified between KG representation learning and KG representation-related tasks.
Next, we will introduce the technical details of Relphormer for KG representation learning.

\subsection{Contextual Sub-Graph Sampling with Triple2Seq}
\label{sec:Triple2Seq}

As stated previously, KG usually contains massive relational facts and thus whole relational graph can not be directly fed into the transformer .
To alleviate such limitations of full-graph-based Transformers, inspired by \cite{Gophormer}, we propose Triple2Seq, which utilizes contextualized sub-graphs as input sequences to encode the local structural information.

The contextualized sub-graph $\mathcal{T}_G$  here is defined as a \bizhen{triple set} containing entities and relations.
$\mathcal{T}_G$ consists of a center triple \bizhen{$\mathcal{T}_c$} and its surrounding neighborhood \bizhen{triple} set $\mathcal{T}_{context}$:

\begin{equation}
    \mathcal{T}_G = \mathcal{T}_c \cup \mathcal{T} _{context} 
\end{equation}

Note that the sub-graph sampling process is at a triple level, then we can obtain a sampled neighborhood \bizhen{triple set} $T_{context}$:

\bizhen{
\begin{small}
\begin{equation}
\begin{aligned}
        &  \mathcal{T}_{context} = \{ \mathcal{T} \mid \mathcal{T}_i \in \mathcal{N}  \}
\end{aligned}
\end{equation}
\end{small}
}

where $\mathcal{N}$ is the fixed-size neighborhood triple set of the center triple \bizhen{ $\mathcal{T}_c$ }.
To better capture the local structural feature, we leverage \bizhen{a dynamic sampling strategy} during training and randomly select multiple contextualized sub-graphs for the same center triple in each \bizhen{epoch}.
We empirically set the sample distribution as $\mathbb{E}_{\mathcal{T}_c \sim P} \propto D$, where $\propto D$ is the degree of triple.

After the procedure of Triple2Seq, we can obtain the contextualized sub-graph $\mathcal{T}_G$.
Meanwhile, the local structure of the contextualized sub-graph is preserved in adjacency matrix $\mathbf{A}_{G}$.
We notice that entity-relation pair information is quite an essential signal for KGs, which proves the effectiveness such as the hierarchical Transformer framework of HittER \cite{HittER}.
Thus, we represent entity-relation pairs as plain text and regard relations as the special nodes in the contextualized sub-graph.
For example, the contextualized sub-graph is converted to a \bizhen{sequence} $ \{ v_1, v_2, ..., v_i \} $, which includes the nodes of center triple \bizhen{$\mathcal{T}_c$}.

In this way, we can obtain node-pair information, including entity-relation, entity-entity and relation-relation pairs interaction.
Another advantage is that the relation node can be seen as a special node.
Since the number of relations in the knowledge graph is much smaller than that of entities, so it can maintain global semantic information among the contextualized sub-graphs sampled by Triple2Seq.
Notably, we also add a \bizhen{global node} to explicitly preserve the global information.
The global node plays a similar role as \texttt{[CLS]} token in natural language pre-training models.
We link the global node with nodes in the contextualized sub-graphs via a learnable virtual distance or fixed distance.
So we obtain the final input sequence with a special virtual node to construct node features.

\begin{remark}
Note that with Triple2Seq, which dynamically samples contextualized sub-graphs to construct input sequences, Transformers can be easily applied to large knowledge graphs.
However, our approach focuses on heterogeneous graphs and regards edges (relation) as special nodes in contextualized sub-graphs for sequential modeling.
Besides, the sampling process can also be viewed as a data augmentation operator  which boosts the performance.
\end{remark}

\subsection{Structure-enhanced Self-attention}
\label{sec:Structural_bias}

Note that the structural information with sequential input may be lost due to the fully-connected nature of the attention mechanism; we propose structure-enhanced self-attention to preserve vital structure information in contextualized sub-graphs.
We utilize attention bias to capture the structural information between node pairs in the contextualized sub-graph.
The attention bias is denoted as $ \phi(i, j)$, which is a bias term between node $v_i$ and node $v_j$.

\begin{equation}
\begin{aligned}
    & a_{ij}=\frac{(\boldsymbol{h}_{i}\mathbf{W}_{Q})(\boldsymbol{h}_{j}\mathbf{W}_{K})}{\sqrt{d}}+\phi{(i,j)} \\
    & \phi{(i,j)} = f_{structure}(\widetilde{\mathbf{A}}^1, \widetilde{\mathbf{A}}^2, ..., \widetilde{\mathbf{A}}^m) \\
\end{aligned}
\end{equation}

where $\widetilde{\mathbf{A}}$ refers to the normalized adjacency matrix, 
{$\mathbf{W}_{Q}$ and $\mathbf{W}_{K}$ are  Query-Key matrix in the transformer module}. 
$\boldsymbol{h}_{i}$, $\boldsymbol{h}_{j}$ denote the hidden representations, {and $d$ is the hidden dimension.} 
The structure encoder $f_{structure}$ is a linear layer with the  $\widetilde{\mathbf{A}}^m$ as the input, where $m$ is a hyper-parameter.
$\widetilde{\mathbf{A}}^m$ ($\widetilde{\mathbf{A}}$ to the $m$-th power) refers the reachable relevance by taking $m$ steps from one node to the other node.
{We concatenate the adjacency matrix as overall structural information.}

For some dense KGs, too many contextualized sub-graphs for the same center triple may cause inconsistency during training.
Thus, we leverage contextual contrastive strategy during the dynamic sampling to overcome the instability.
We use the contextualized sub-graphs of the same triple in different epochs to enforce the model to conduct similar predictions.
We encode the input sequence of a contextualized sub-graph and take the hidden vector $\boldsymbol{h}_{mask}$ as contextual representation $\boldsymbol{c}_t$ at current epoch $t$, and $\boldsymbol{c}_{t-1}$ at last epoch $t-1$. The goal is to minimize the differences between different sub-graphs; we get the contextual loss $\mathcal{L}_{contextual}$ as:

\begin{small}
\begin{equation}
\begin{aligned}
  \mathcal{L}_{contextual} =  & - log \frac{exp(sim(\boldsymbol{c}_{t}, \boldsymbol{c}_{t-1})/\tau )}{exp(sim(\boldsymbol{c}_t,\boldsymbol{c}_{t-1})/\tau) + \sum_{j}{ exp(sim(\boldsymbol{c}_t, \boldsymbol{c}_{j}) /\tau )}}
\end{aligned}
\end{equation}
\end{small}

where $\tau$ denotes a temperature parameter and $sim(\boldsymbol{c}_t, \boldsymbol{c}_{t-1})$ is the cosine similarity $\frac{{\boldsymbol{c}}_t^{{T}}{\boldsymbol{c}}_{t-1}}{\|{c}_t\| \cdot \|{\boldsymbol{c}}_{t-1} \|}$.
$\boldsymbol{c}_t$ is the hidden state representation at epoch $t$, which belongs to different center triples.

\begin{remark}
It should be noted that our structure-enhanced Transformer is model-agnostic and, therefore, orthogonal to existing approaches, which injects semantic and structural information into the Transformer architecture.
In contrast to \cite{transformers} where attention operations are only performed between nodes with literal edges in the original graph, structure-enhanced Transformer offers the flexibility in leveraging the local contextualized sub-graph structure and influence from the semantic features, which is convenient for information exchange between similar nodes in the local graph structure. 
\end{remark}

\subsection{Masked Knowledge Modeling}
\label{sec:MKM}
In this section, we introduce the proposed masked knowledge modeling for KG representation.
Typically, KG completion aims to learn a mapping $f:\mathcal{T}_M, \mathbf{A}_G \to Y$.
Inspired by masked language modeling, we randomly mask specific tokens of the input sequences and then predict those masked tokens.

Given an input contextualized sub-graph node sequence $\mathcal{T}_G$, we randomly mask tokens in the center triple.
Concretely, the masked center triple will be head or tail entity for relation prediction task, respectively.
We denote the candidates set as $Y$, and the task of masked knowledge modeling is to predict the missing part of the triple $\mathcal{T}$ given the masked node sequence $\mathcal{T_{M}}$ and contextualized sub-graph structure $\mathbf{A}_{G}$ as:

\begin{equation}
\begin{aligned}
    & \mathcal{T}_{M} =MASK(\mathcal{T}_{G}) \\
    & {Relphormer}(\mathcal{T}_{M}, {\mathbf{A}_{G}}) \to {Y}\\
\end{aligned}
\end{equation}
where $Y \in \mathbb{R}^{ \left |  \mathcal{E}  \right | + \left |  \mathcal{R}  \right | }$.
Specifically, we randomly mask only one token for a sequence to better integrate the contextual information due to the unique structure of the contextualized sub-graph.

Intuitively, masked knowledge modeling is different from previous translational distance methods, which can avoid the defects brought by score-function-based methods.
However, such as strategy may cause a serious label leakage problem for link prediction if we simultaneously sample the head and tail entity's neighborhood nodes.
Note that during training, the structure of the predicted mask token should be unknown.
To handle the label leakage issue and bridge the gap between training and testing, we {remove the context nodes of target entities to ensure fair comparison}.

\begin{remark}
The advancement of empirical results (See section \ref{sec:exp}) illustrates that masked knowledge modeling may be a parametric score function approximator, which can automatically find a suitable optimization target for better link prediction.

\begin{proof_}
(score function approximator) Let $~\mathcal{T}_{M} $ be the masked triplet, $\textbf{h}  \in  \mathbb{R}^d $ is the masked head derived from multi-head attention layers in Relphormer $\mathcal{M}(\theta)$. 
The vocabulary token embedding is $ W  \in \mathbb{R}^{d \times N} $, where $ N = \left |  \mathcal{E}  \right | + \left |  \mathcal{R}  \right |$. 

If the $~\mathcal{T}_{M}$ is the triplet $(v_{s}, v_{p}, \texttt{[MASK]})$, where the tail entity is masked.
We define the $g(\cdot)$ function as the multi-head attention layer modules and $V_{object} \subset W $ the candidate tail entities embeddings. The output logits are $ \texttt{sigmoid}(W \textbf{h})$, approximately equal to $\texttt{sigmoid}(V_{object} \textbf{h})$.
Then we get the final logits:

\begin{equation}
\begin{aligned}
    \texttt{sigmoid}\sum^{ \left | \mathcal{E} \right | }{v_{{object}_i}  g(v_{s}, v_{p}, \texttt{[MASK]})}
\end{aligned}
\end{equation}

We choose one term $ f(v_{s}, v_{p}, v_{{object}_i})  $. 
and use $f(\cdot) \approx v_{{object}_i}g(\cdot)$, which acts as a score function role.
Therefore masked knowledge modeling may be the score function approximator.
\end{proof_}
\end{remark}

\subsection{Training and Inference}
\label{sec:downtream_task}
\textbf{Training Relphormer for KG representation learning.}
The overall Relphormer optimization procedure is illustrated in \bizhen{Algorithm \ref{alg1}}.
During training, we jointly optimize masked knowledge loss and contextual contrastive constrained objects. 
The $\mathcal{L}_{contextual}$ can be viewed as a constraint term to the whole loss $\mathcal{L}_{all}$ as follows:

\begin{equation}
     \mathcal{L}_{all} = \mathcal{L}_{MKM} +  \lambda \mathcal{L}_{contextual}
\end{equation}
where $\lambda$ is a hyper-parameter, $\mathcal{L}_{MKM}$ and $\mathcal{L}_{contextual}$ are mask knowledge and contextual loss.

\textbf{KG Completion}
During inference, we use a multi-sampling strategy for testing following \cite{Gophormer}, which can enhance the stability of prediction:

\begin{equation}
\begin{aligned}
     \widetilde{\boldsymbol{y}} = \frac{1}{K} \sum_{k} \boldsymbol{y}_k \\
\end{aligned}
\end{equation}

where $\boldsymbol{y}_k \in \mathbb{R}^{ \left |  V \right | \times 1}$ refers to the predicted result of one contextualized sub-graph, and $K$ denotes the number of sampled  sub-graphs.


  
  



\begin{algorithm}[t]
  \caption{{Relphormer Training Algorithm}}
  \label{alg1}

      \SetAlgoLined

      {  
          $\mathcal{M}(\theta)$: model of Relphormer, $\alpha$: learning rate \;
          $h \in \theta$: initial parameters of Relphormer \;
          $\mathcal{D}$: training data set, a list of triples\;
      }

      \BlankLine      
        { 
              Initialize pre-trained parameters of $h$\;    
        }
      
      \While{$\mathcal{M}(\theta)$ not converged}{
        \For{each $triple ~  \mathcal{T} \in \mathcal{D}$}{
          \tcp{Obtain the contextualized sub-graph sequence}
          $\mathcal{T}_G,\mathbf{A}_G \gets Triple2Seq(\mathcal{T})$ \;
          \tcp{Mask the contextualized sub-graph sequence} 
          $\mathcal{T}_{M} \gets MASK(\mathcal{T}_G)$ \;
          \tcp{Masked knowledge prediction} 
          $\mathcal{L} \gets \mathcal{M}(\mathcal{T}_{M}, \mathbf{A}_{G})$  \;
          \tcp{Calculate loss and update parameters $\theta$ using back-propagation}
          $\theta \gets Update(\theta, \mathcal{L}, \alpha)$
        }
      }  
  
\end{algorithm}

\textbf{Question Answering \& Recommendation.}
We introduce the fine-tuning strategy of Relphormer (pre-trained with KG) for KG-based downstream tasks including question answering and recommendation, which is formulated as a mapping:

\begin{equation}
    f:\mathcal{Q}_M, \mathcal{M}(\theta) \to Y
\end{equation}

where $\mathcal{Q}_M$ is the masked query and $\mathcal{M}(\theta)$ is the pre-trained KG transformer.
For different downstream tasks, the masked query $\mathcal{Q}_M$ can be defined in different forms.
For example, the masked query for QA is defined by $[question~tokens; \texttt{[MASK]}]$ (predicts the answer entities in the KG),and  $[items~tokens; \texttt{[MASK]}]$ for the recommendation task.

\begin{table}[!t]

    \caption{\textbf{Inference efficiency comparison.} 
    $ \left | \mathcal{E} \right |$, $ \left | \mathcal{R} \right |$ and $ \left | \mathcal{G} \right |$ are numbers of  entities, relations, inference samples in the graph respectively.
    $k$ is the length of input sequence.
    The time refers to the speed of inference time given a single (entity, relation) on FB15K-237.
    } 

    \centering
        \scalebox{0.85}{
    \begin{tabular}{llll}
    \toprule
    \textbf{Inference} & \textbf{Method} & \textbf{Complexity} & \textbf{Speed up}\\ 
    \midrule
        
    \multirow{2}{*}{Triple}
        & KG-BERT 
        & $\mathcal{O}( k (\left | \mathcal{E}  \right |  +  \left |  \mathcal{R}  \right |  ))$ & $\sim \left | \mathcal{E}  \right | $ 
        \\ 
        
        & Relphormer 
        & $\mathcal{O}(k+1)$ 
        &  
        \\
    \midrule
    \multirow{2}{*}{Graph} 
        & KG-BERT 
        & $\mathcal{O}( k  (\left | \mathcal{E}  \right | + \left |  \mathcal{R}  \right |  ) \times \mathcal{G} )$ 
        &  $\sim \left | \mathcal{E}  \right | $ \\   
        & Relphormer & $\mathcal{O}((k + 1) \times \mathcal{G}) $ 
        & \\
    
    \bottomrule

    \end{tabular}
    }
   
     \label{tab:inference_efficency}

\end{table}

\textbf{Model Time Complexity Analysis}
\label{sec:time}
We further analyze the inference speed between Relphormer and other Transformer models.
From Table \ref{tab:inference_efficency}, we notice that the approach with masked knowledge modeling obtains faster inference speed than KG-BERT.
We note that KG-BERT uses the Transformer as its encoder and the translational distance score function as the decoder.
Since KG-BERT has to repeatedly calculate the scores of candidates thus, it is time-consuming during inference.
However, Relphormer leverages the masked knowledge modeling strategy, and the model infers the targets by predicting the masked entities or relations. 
Although the Triple2Seq procedure costs some time, Relphormer is still much faster than KG-BERT in inference.

\section{Experiments}
\label{sec:exp}
Extensive experiments are conducted to evaluate the performance
of the Relphormer by answering the following research questions:
\begin{itemize}
    \item \textbf{RQ1}: How does our Relphormer perform when competing with baselines of KG representation for link prediction?
    \item \textbf{RQ2}: How does the KG representations from our Relphormer benefit the downstream tasks including question answering and recommendation?
    \item \textbf{RQ3}: How do different key modules in our Relphormer framework contribute to the overall performance?
    \item \textbf{RQ4}: How effective is the proposed Relphormer model in addressing heterogeneity KG structure and semantic textual description?
\end{itemize}

\subsection{Experimental Settings}
\subsubsection{\textbf{Datasets.}}
We evaluate the proposed Relphormer on six popular benchmark datasets.
For the Knowledge Graph completion (KGC) task, we use three publicly available datasets for evaluation: \textbf{WN18RR} \cite{WN18RR},  \textbf{FB15K-237} \cite{FB15K237} and \textbf{UMLS} \cite{UMLS}. 
WN18RR is a subset of the WordNet and contains a lexical knowledge graph for English. 
FB15K-237 is a subset of the Freebase and is constructed by limiting the set of relations in FB15K.
UMLS is a small dataset that contains medical semantic entities and relations.
For the QA task, we use two publicly available datasets for evaluation: \textbf{FreeBaseQA} \cite{FreebaseQA}, \textbf{WebQuestionSP} \cite{WebQuestionsSP}.
FreebaseQA is a data set for open-domain QA over the Freebase knowledge graph.
WebQuestionSP contains semantic parses for the questions from WebQuestions that are answerable using Freebase.
For the recommendation task, we adopt the well-established versions \textbf{ML-20m} in \textbf{MovieLens} \cite{MovieLens}  dataset and obtain textual descriptions of movies from Freebase.

\subsubsection{\textbf{Evaluation Protocols}.}
For KG completion, we follow the settings in \cite{HittER} and evaluate the performance of KG representation models under \textit{Filter-} and \textit{Full-} settings. 
We use \textbf{MR} (Mean Rank), \textbf{MRR} (Mean Reciprocal Rank) and \textbf{Hit@1, 3, 10} (Hit Ratio values) as the main evaluation metrics. 
For the QA task, we follow \cite{HittER} to pre-train our Relphormer in FB15k237, and then use the pre-trained KG model for fine-tuning.
We utilize accuracy as main evaluation metric.
For the recommendation task, we pre-train our Relphormer in the item-item KG and then leverage those KG representations as augmentation.
We utilize the MRR as main evaluation metric.

\subsubsection{\textbf{Baselines for Comparison.}}
We compare our Rephormer with various baselines, namly:
\begin{itemize}

   \item  \textbf{KG Completion.}
We compare Relphormer with several baseline models to demonstrate the effectiveness of our proposed approach.
Firstly, we choose the tranlational-based models, such as TransE \cite{TransE} and RotatE  \cite{RotatE}.
We regard R-GCN \cite{RGCN} as a translational distance model because  R-GCN still uses the score function as its decoder.
Further, we compare our Relphormer with models based on Transformer architecture such as KG-BERT \cite{KG-BERT}, HittER \cite{HittER}, and StAR \cite{StAR}.

   \item  \textbf{Question Answering.}
We use BERT as the backbone in two datasets.
Then we compare our Relphormer with BERT, where two models of (HittER \cite{HittER} and Relphormer) are first pre-trained in the FB15K-237 dataset and then injected into the BERT module.

   \item  \textbf{Recommendation.}
We use BERT4Rec \cite{BERT4Rec}  as baselines  in ML-20m since it leverages the same Transformer architecture which is flexible to inject embeddings from Relphormer\footnote{Our Relphormer can also be integrated into other recommendation approaches, but due to the page limit we only provide one evaluation and leave this for future works.}.
Note that our model should leverage the entity descriptions, we utilize the raw data of ML-20m and Freebase offered by KB4Rec \cite{Zhao-DI-2019} which contains textual descriptions of movies in ML-20m. 
Then we compare our Relphormer with BERT4REC and KG-BERT, where two models of KG-BERT and Relphormer are first pre-trained in the KG and then injected into the BERT module.

\end{itemize}

\begin{table*}[htbp!]
        
        \caption{Results of the link prediction on FB15K-237, WN18RR and UMLS. The bold numbers denote the best results, while the underlined ones are the second-best performance.}
        
          \centering
            \scalebox{0.85}{
                \begin{tabular}{c|rrr|rrr|rr}
           \toprule
                \multirow{2}{*}{Model}      
                  & \multicolumn{3}{c|}{\textbf{WN18RR}}
                  & \multicolumn{3}{c|}{\textbf{FB15K-237}} 
                  & \multicolumn{2}{c}{\textbf{UMLS}} \\
            \cmidrule{2-9} 
                  & \multicolumn{1}{c}{Hits@1 } 
                  & \multicolumn{1}{c}{Hits@10 } 
                   & \multicolumn{1}{c|}{MRR } 
                 & \multicolumn{1}{c}{Hits@1 } 
                  & \multicolumn{1}{c}{Hits@10 } 
                   & \multicolumn{1}{c|}{MRR } 
                  & \multicolumn{1}{c}{Hits@10 } 
                   & \multicolumn{1}{c}{MR } 
                   \\
                   
            \midrule
            
            \multicolumn{1}{c}{\textbf{Translational distance models}} 
            & \multicolumn{1}{c}{} 
            & \multicolumn{1}{c}{} 
            & \multicolumn{1}{c}{}
            
            & \multicolumn{1}{c}{} 
            & \multicolumn{1}{c}{} 
            & \multicolumn{1}{c}{}
            
           
          &\multicolumn{1}{c}{}
           & \multicolumn{1}{c}{} 
           
           \\     
           
            \midrule
        
            \multicolumn{1}{c|}{TransE~\cite{TransE}} 
            & \multicolumn{1}{c}{0.061}
            & \multicolumn{1}{c}{0.522}
            & \multicolumn{1}{c|}{0.232}
        
           &\multicolumn{1}{c}{0.218} 
           &\multicolumn{1}{c}{0.495} 
           & \multicolumn{1}{c|}{0.310} 
           

           &\multicolumn{1}{c}{0.989} 
           & \multicolumn{1}{c}{1.84} \\
           
           \multicolumn{1}{c|}{R-GCN~\cite{RGCN}} 
            & \multicolumn{1}{c}{0.080} 
            & \multicolumn{1}{c}{0.207} 
            & \multicolumn{1}{c|}{0.123}
            
           &\multicolumn{1}{c}{0.100} 
           &\multicolumn{1}{c}{0.300}
           & \multicolumn{1}{c|}{0.164}
           

           &\multicolumn{1}{c}{-} 
           & \multicolumn{1}{c}{-} \\
            
            \multicolumn{1}{c|}{DistMult~\cite{DistMult}}
            & \multicolumn{1}{c}{0.412} 
            & \multicolumn{1}{c}{0.504}
            & \multicolumn{1}{c|}{0.444}
              
            &\multicolumn{1}{c}{0.199} 
            &\multicolumn{1}{c}{0.446}
            & \multicolumn{1}{c|}{0.281} 
            

           &\multicolumn{1}{c}{0.846} 
           & \multicolumn{1}{c}{5.52} \\
            
            \multicolumn{1}{c|}{ConvE~\cite{ConvE}}
            & \multicolumn{1}{c}{0.419} 
            & \multicolumn{1}{c}{0.531}
            & \multicolumn{1}{c|}{0.456}
              
            &\multicolumn{1}{c}{0.225} 
            &\multicolumn{1}{c}{0.497}
            & \multicolumn{1}{c|}{0.312} 
            
            
           &\multicolumn{1}{c}{0.990} 
           & \multicolumn{1}{c}{1.51} \\
           
            \multicolumn{1}{c|}{ComplEx~\cite{ComplEx}}
            &\multicolumn{1}{c}{0.409}
            & \multicolumn{1}{c}{0.530}
            & \multicolumn{1}{c|}{0.449}
         
            & \multicolumn{1}{c}{0.194}
            & \multicolumn{1}{c}{0.450}
            & \multicolumn{1}{c|}{0.278}
            

           &\multicolumn{1}{c}{0.967} 
           & \multicolumn{1}{c}{2.59} \\
           
            \multicolumn{1}{c|}{RotatE~\cite{RotatE}}
            & \multicolumn{1}{c}{0.428} 
            & \multicolumn{1}{c}{0.571}
            & \multicolumn{1}{c|}{{0.476}}
              
            &\multicolumn{1}{c}{0.241} 
            &\multicolumn{1}{c}{\underline{0.533}}
            & \multicolumn{1}{c|}{{0.338}} 
            

           &\multicolumn{1}{c}{-} 
           & \multicolumn{1}{c}{-} \\
            
            \multicolumn{1}{c|}{QuatE~\cite{QuatE}}
            & \multicolumn{1}{c}{\underline{0.436}} 
            & \multicolumn{1}{c}{0.564}
            & \multicolumn{1}{c|}{{0.481}}
              
            &\multicolumn{1}{c}{0.221} 
            &\multicolumn{1}{c}{0.495}
            & \multicolumn{1}{c|}{0.311} 
            

           &\multicolumn{1}{c}{-} 
           & \multicolumn{1}{c}{-} \\
           
          \midrule
          
          \multicolumn{1}{c}{\textbf{Transformer-based models}} 
            & \multicolumn{1}{c}{} 
            & \multicolumn{1}{c}{} 
            & \multicolumn{1}{c}{}
            
           &\multicolumn{1}{c}{} 
           &\multicolumn{1}{c}{}
           & \multicolumn{1}{c}{} 
           

           &\multicolumn{1}{c}{} 
           & \multicolumn{1}{c}{} \\
           
           \midrule
             \multicolumn{1}{c|}{KG-BERT~\cite{KG-BERT}}
             & \multicolumn{1}{c}{0.041} 
             & \multicolumn{1}{c}{0.524} 
             & \multicolumn{1}{c|}{0.216}
             
             &\multicolumn{1}{c}{-} 
           &\multicolumn{1}{c}{0.420} 
           & \multicolumn{1}{c|}{-}
           

           &\multicolumn{1}{c}{0.990} 
           & \multicolumn{1}{c}{\textbf{1.47}} \\
             \multicolumn{1}{c|}{HittER~\cite{HittER}} 
             &\multicolumn{1}{c}{\underline{0.436}} 
          &\multicolumn{1}{c}{0.579} 
          & \multicolumn{1}{c|}{\underline{0.485}}
           
            &\multicolumn{1}{c}{\underline{0.279}} 
          &\multicolumn{1}{c}{\textbf{0.558}}
          & \multicolumn{1}{c|}{\textbf{0.373}} 
          

           &\multicolumn{1}{c}{-} 
           & \multicolumn{1}{c}{-} \\

             \multicolumn{1}{c|}{StAR~\cite{StAR}}
             & \multicolumn{1}{c}{0.243} 
             & \multicolumn{1}{c}{\textbf{0.709}} 
             & \multicolumn{1}{c|}{0.401}
             
             &\multicolumn{1}{c}{0.205} 
           &\multicolumn{1}{c}{0.482} 
           & \multicolumn{1}{c|}{0.296}
           

           &\multicolumn{1}{c}{\underline{0.991}} 
           & \multicolumn{1}{c}{\underline{1.49}} \\
          
             \midrule
        
            \multicolumn{1}{c|}{{{Relphormer}}} 
            & \multicolumn{1}{c}{\textbf{0.448}}
            & \multicolumn{1}{c}{\underline{0.591}} 
            & \multicolumn{1}{c|}{\textbf{0.495}}
            
           &\multicolumn{1}{c}{\textbf{0.314}} 
           &\multicolumn{1}{c}{{{0.481}}}
           & \multicolumn{1}{c|}{\underline{0.371}}
           

           &\multicolumn{1}{c}{\textbf{0.992}} 
           & \multicolumn{1}{c}{1.54} \\
        
            \bottomrule
            \end{tabular}
            
            }

        \label{tab:link_prediction}
        \end{table*}
\begin{table}[htbp!]
        
        \caption{Results of relation prediction in FB15K-237, WIN18RR. 
            \label{tab:relation_prediction}
        }
        
        \centering
            \scalebox{0.85}{
        \begin{tabular}{c|cccccc}
        \toprule
        \multirow{2}{*}{Model}               
        & \multicolumn{3}{c}{\textbf{WN18RR} }  & \multicolumn{3}{c}{\textbf{FB15K-237} } \\ 
        \cmidrule{2-7} 
        & MRR $\uparrow$ & Hit@1 & Hit@3 & MRR $\uparrow$ & Hit@1 & Hit@3  \\ 
        \midrule
        TransE \cite{TransE} & 0.784 & 0.669 & 0.870 & {0.966} & {0.946} & 0.984\\
        ComplEx \cite{ComplEx}  & 0.840 & 0.777 & 0.880 & 0.924 & 0.879 & 0.970 \\
        DistMult \cite{DistMult} & 0.847 & {0.787} & 0.891 & 0.875 & 0.806 & 0.936 \\
        RotatE \cite{RotatE} & 0.799 & 0.735 & 0.823 & \textbf{0.970} & \textbf{0.951} & 0.980 \\
        DRUM \cite{DRUM} & {0.854} & 0.778 & {0.912} & 0.959 & 0.905 & {0.958} \\
        \midrule
        Relphormer & \textbf{0.897} & \textbf{0.827} & \textbf{0.967} & 0.958 & 0.930 & \textbf{0.986} \\
        \bottomrule
        \end{tabular}
            }

\end{table}

\subsection{Implementation Details}
\label{appendix:implementation_details}

We initialize Relphormer with BERT-base.
We utilize Pytorch to conduct experiments with 1 Nvidia 3090 GPU. 
Early stopping is adopted to reduce over-fitting on the training set.
We use BERT-base to encode textural descriptions for semantic  features.
The number of contextualized sub-graph nodes is chosen from set $\{8, 16, 32, 64\}$ by tuning on the development set.

\subsection{KG Completion Results (RQ1)}
        \subsubsection{\textbf{Entity Prediction}}
        From Table \ref{tab:link_prediction}, we observe that our proposed approach can achieve competitive performance on all datasets compared with baselines.
        Relphormer achieves the best performance on Hits@1 and MRR metrics and yields the second-best performance on Hits@10 in WN18RR. 
        Compared to the previous SOTA translational distance models, such as QuatE, our method has improvements in all metrics.
        We notice that Relphormer is superior to SOTA Transformer-based model HittER in WN18RR.
        In the FB15K-237 dataset, we also find that Relphormer outperforms most translational distance models.
        Compared with transformer-based models, we find that Relphormer has the best performance on hits@1, which is better than KG-BERT, StAR and HittER.
        HittER advances the performance in FB15K-237 since they explicitly utilize more transformer architectures, and our proposed method still obtains comparable performance.
        Besides, we notice that Relphormer obtains the best performance in UMLS, especially on Hits@10.
        The excellent performance of Relphormer proves that our relational transformer framework is helpful for knowledge completion tasks.

    \subsubsection{\textbf{Relation Prediction}}
        From Table \ref{tab:relation_prediction}, we observe that our Relphormer can obtain competitive performance compared with  baselines.
        In the WN18RR dataset, Relphormer can already outperform all baselines, which demonstrates the excellent performance of our approach for relation prediction.
        Compared to TransE, our method improves 15.8\% on Hits@1, 9.7\%.
        In the FB15K-237 dataset, the improvement of Relphormer is significant on Hits@3.
        Relphormer performs better than DistMult but worse than RotatE {on MRR and Hit@1}.
        {
        It is because the relation prediction task is relatively simpler compared to entity prediction, but our method is still comparable to the SOTA methods.    
        }

\begin{figure*}
    \centering
    \includegraphics[width=1.0\textwidth]{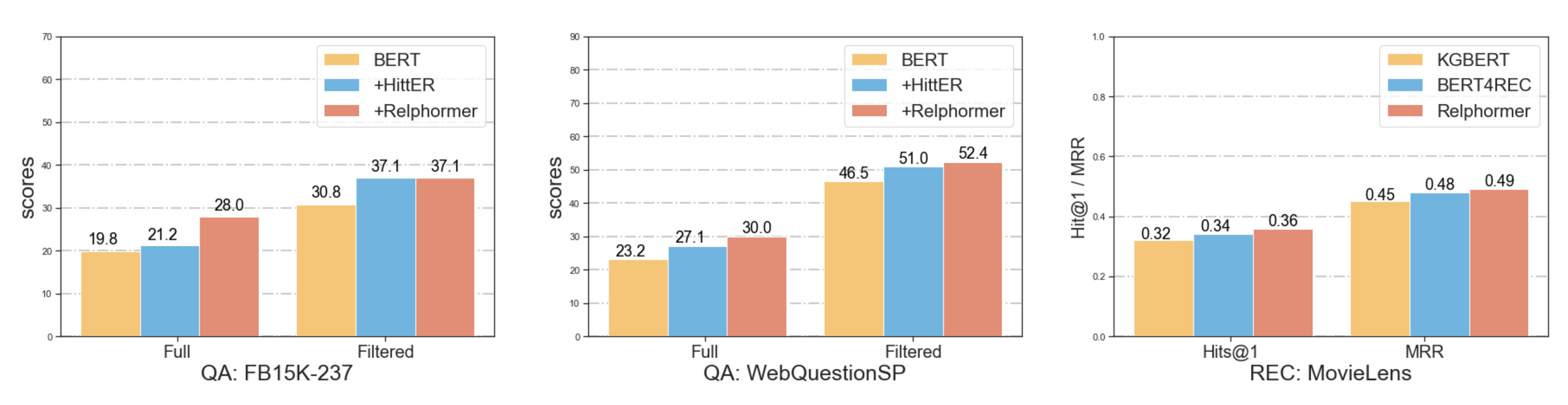}
        \caption{
    \bizhen{Results of KG-based tasks.
    Left: results of QA on {FreebaseQA}.
    Middle: results of QA on WebQuestionSP.
    Right: results of Recommendation on Movielens.
    }
    }
    \label{fig:qa_rec}
\end{figure*}

\begin{figure*}
    \centering
    \includegraphics[width=1.0\textwidth]{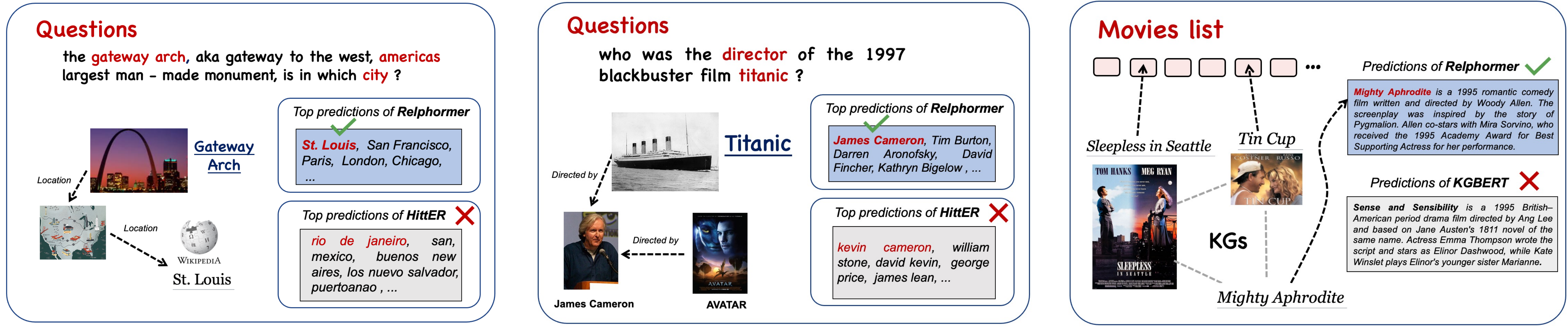}
    \caption{Case study of KG-based tasks.
    Left: Case of QA task on {FreebaseQA}.
    Right: Case of Recommendation task on Movielens.
    }
    \label{fig:case_qa_rec}
\end{figure*}

\subsection{Question Answering and Recommendation Results (RQ2)}

For the question answering task, we find that Relphormer obtains the best performance on FreebaseQA and WebQuestionSP in Figure \ref{fig:qa_rec}.
Compared to HittER, Relphormer improves 6.8\% in the \textit{Full} setting in the FreebaseQA dataset. 
Meanwhile, Relphormer improves 2.9\% and 1.4\% in the 
\textit{Full}  and \textit{Filter} settings on WebQuestionSP.
Note that Relphormer can be easily initialized with BERT and is optimized with masked knowledge modeling, so it is efficient to inject \bizhen{pre-trained representations} with Relphormer for the QA task, leading to better performance.
We find that Relphormer is much more efficient.
For some KG representation models such as HittER, they need to design complicated integration strategies to enhance the QA task.
One straightforward way is to inject the pre-trained representation into an extra QA model. 
However, due to the discrepancy between the pre-trained KG model and the downstream model, it is hard to verify the effectiveness.

For hard samples with abundant textural and structural information on FreebaseQA (Figure \ref{fig:case_qa_rec}), we note that Relphormer can learn the explicit and implicit correlation between different words or entities.
In the first sample, we need to find the location of "gateway arch".
From this case, we see that our method does well in employing existing knowledge to help answer the question.
But the baseline model fails to use textual and structural information from the knowledge graph.

For the recommendation task, Figure  \ref{fig:qa_rec}  shows that Relphormer performs better than all baseline models in the recommendation task.
Compared to BERT4REC, Relphormer improves 2\% on Hits@1 and 1\% on MRR.
Moreover, Relphormer outperforms KG-BERT, which is implemented by each node's BERT embeddings calculation and aggregations.
As shown in Figure \ref{fig:case_qa_rec}, 
A specific user has watched a long list of movies, the goal here is to predict the next movie he/she will choose to watch.
Among those movies, \textit{Sleepless in Seattle} and \textit{Tin Cup} are highly correlated, because both their themes are about romance and comedy.
Meanwhile,  movie \textit{Mighty Aphrodite} in the additional KGs is connected to \textit{Sleepless in Seattle} and \textit{Tin Cup} for the same reason.
Obviously, the potential node relevance in the KGs is beneficial for the movie recommendation task. 
For these samples, our method will learn the deep correlation and yield better performance than baseline models.
Overall, we demonstrate that KG representations with Relphromer can perform better intrinsic evaluation performance with link prediction as well as promote KG-based downstream tasks of question answering and recommendation with well-learned knowledge representations.

\begin{figure*}
    \centering

    \includegraphics[width=0.9\textwidth]{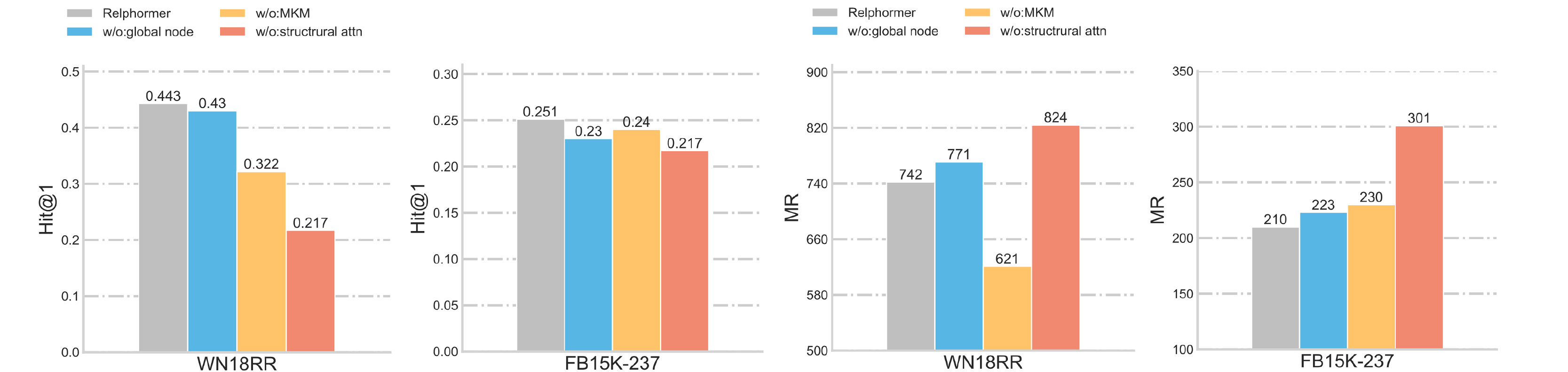}

    \caption{
    Ablation study on WN18RR and FB15K-237.
    Left: results of different model variants on Hits@1.
    Right: results of different model variants on Mean Rank.
    }
    
\label{fig:others}
\end{figure*}

\subsection{Ablation Study of Relphormer (RQ3)}

\subsubsection{\textbf{Optimization object}}
In the knowledge graph, there are relation-specific patterns, and some approaches can not solve patterns like 1-N, N-1 and N-N relations.
For example, given a specific entity-relation pair $(h, r)$, there usually exists more than one tail entity as labels.
It seems to be more significant to investigate whether different optimization objects have an impact on Relphormer.
Specifically, we conduct an ablation study without masked knowledge modeling (\textit{w/o MKM}) but using the negative log-likelihood loss instead.
From Table \ref{fig:others}, we notice that the model with MKM can yield better performance on Hit@1 on both datasets; however, it fails to achieve advancement in MR on WN18RR. 
This may be because WN18RR lacks enough structural features; thus, NLL-based optimization for ranking may be more advantageous.

\begin{figure}
    \centering  

    \includegraphics[width=0.4\textwidth]{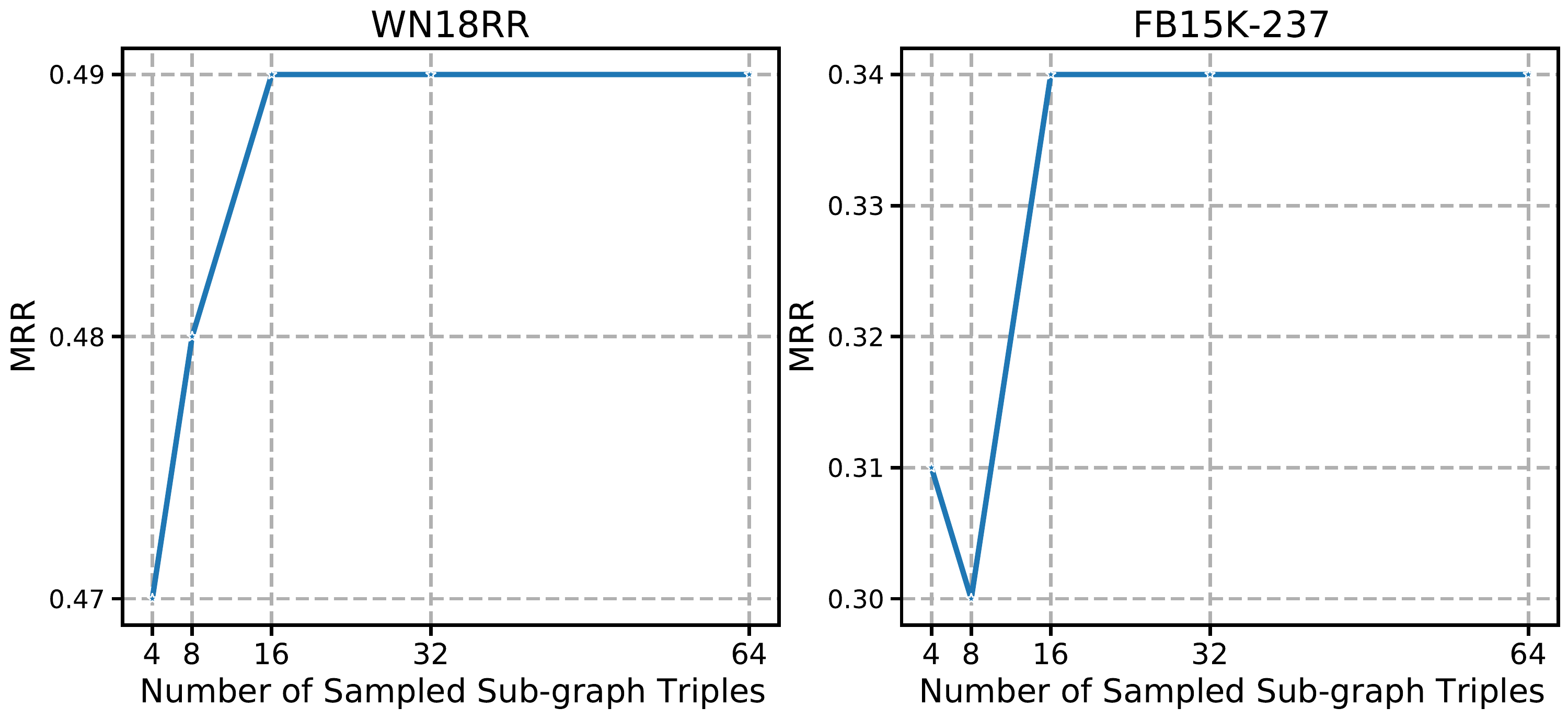}

    \caption{Ablation study on WN18RR and FB15K-237.
    Results of the increasing number of contextualized sub-graph triples.
    }

    \label{fig:context_number}
\end{figure}

\subsubsection{\textbf{Global node}}
We further conduct an ablation study to analyze the impact of the global node.
\textit{w/o global node} refers to the model without the global node.
From Figure \ref{fig:others}, we can see that the model without a global node achieves poorer performance than baselines. 
It demonstrates that the global node might be a more helpful solution to preserving global information.










\begin{table*}[t]
\center
\setlength{\tabcolsep}{3pt}
\caption{
\small Ranking results of target entities for different approaches. 
The first column includes incomplete triples for inference, as well as their labels. 
And the others include the ranking position and whether the link prediction result is right (ranking is not more than 10). 
} 

\scalebox{0.65}{
\begin{tabular}{c|ccc}
\hline
\multicolumn{1}{c|}{\multirow{2}{*}{Incomplete Triple}} & \multicolumn{3}{c}{Positive entity ranking position \& Link prediction result is True or False} \\ \cline{2-4} 
\multicolumn{1}{c|}{}                        & \multicolumn{1}{c|}{\textbf{Relphormer}[Textual Encoding]} & \multicolumn{1}{c|}{\textbf{StAR} [Textual Encoding]} & \textbf{RotatE} [Graph Embedding] \\ \hline
\multicolumn{1}{c|}{\multirow{2}{*}{\begin{tabular}[c]{@{}c@{}}(?  \textit{, /music/genre/artists, Danzig}) \\ $\leftarrow$ \textit{Blues-rock} \end{tabular}}}
& \multicolumn{1}{c|}{\multirow{2}{*}{\begin{tabular}[c]{@{}c@{}}2, \textit{True} \end{tabular}}}
& \multicolumn{1}{c|}{\multirow{2}{*}{\begin{tabular}[c]{@{}c@{}}64, \textit{False} \end{tabular}}}  
& \multirow{2}{*}{\begin{tabular}[c]{@{}c@{}}7, \textit{True} \end{tabular}} \\
                                             & \multicolumn{1}{l|}{}                 & \multicolumn{1}{l|}{}      &        \\ \hline
\multicolumn{1}{c|}{\multirow{2}{*}{\begin{tabular}[c]{@{}c@{}}(\textit{Marquette University, school type, }?) \\ $\leftarrow$ \textit{Society of Jesus-GB} \end{tabular}}}
& \multicolumn{1}{c|}{\multirow{2}{*}{\begin{tabular}[c]{@{}c@{}}3, \textit{True}\\  \end{tabular}}}
& \multicolumn{1}{c|}{\multirow{2}{*}{\begin{tabular}[c]{@{}c@{}}18, \textit{False}\\ \end{tabular}}}  
& \multirow{2}{*}{\begin{tabular}[c]{@{}c@{}}5, \textit{True}\\ \end{tabular}} \\
                                             & \multicolumn{1}{l|}{}                 & \multicolumn{1}{l|}{}      &        \\  [2pt] \hline
\multicolumn{1}{c|}{\multirow{2}{*}{\begin{tabular}[c]{@{}c@{}}(?  \textit{, /music/record label/artist, John 5}) \\ $\leftarrow$ \textit{Interscope Records} \end{tabular}}}
& \multicolumn{1}{c|}{\multirow{2}{*}{\begin{tabular}[c]{@{}c@{}}2, \textit{True} \end{tabular}}}
& \multicolumn{1}{c|}{\multirow{2}{*}{\begin{tabular}[c]{@{}c@{}}40, \textit{False} \end{tabular}}}  
& \multirow{2}{*}{\begin{tabular}[c]{@{}c@{}}10, \textit{True} \end{tabular}} \\
                                             & \multicolumn{1}{l|}{}                 & \multicolumn{1}{l|}{}      &        \\ [3pt] \hline
\end{tabular}
}

\label{tab:Case}
\end{table*}

\subsubsection{\textbf{Case analysis}}

As shown in Table \ref{tab:Case}, we list several triples in the FB15k-237 dataset, where the prediction result obtains obvious improvements when applying our Relphormer.
Relphormer can simultaneously handle the structural and textual features and solve the heterogeneity problem of both entity and relation levels.
As a result, our approach still keeps excellent performance for some difficult samples with plenty of heterogeneous features.

Specially, we know that RotatE is a simple but effective method for embedding entities and relations. 
It can model complex relations in a unified framework, which is achieved by representing relations as complex numbers in a high-dimensional space.
But RotatE is lack of ability for encoding textual information.
To handle both structural and textual information, additional methods such as incorporating textual embeddings or other neural networks must be used in conjunction with RotatE.
By contrast, our approach maintains good performance even for difficult samples with many heterogeneous features.
StAR is a structure-augmented text representation learning method.
However, it has been observed in Table \ref{tab:Case}  that on the WN18RR dataset, StAR does not achieve ideal performance even when incorporating textual information.
The reason might be the defect in the method itself for handling both textural and structural features.

\subsection{Benefits of Relphormer in Addressing the Structure and Semantic Issues (RQ4)}

\subsubsection{\textbf{The number of sampled contextualized sub-graph triples.}}
We investigate the performance of our model with different numbers of sampled contextualized sub-graph triples.
We change the number of triples from 4 to 64 and report the results of entity prediction in WN18RR and FB15K-237 in Figure \ref{fig:context_number}.
When the number is smaller, a few contextual nodes are sampled surrounding the center triple. 
It is evident to see that increasing contextualized sub-graphs give a boost in performance in both two datasets, which indicates that our proposed method can effectively encode contextual information.
However, when the number is too large, the performance of the model will no longer increase.
We argue that although neighborhood information is beneficial, too many unrelated nodes will bring out unnecessary noisy signals, thus, affecting the performance.
Note that those low-quality or irrelevant nodes may lead to negative contextual information infusion, which is detrimental to the performance. 
It is intuitive to investigate selective contextual information integration.

\begin{figure}
    \centering
    \includegraphics[width=0.5\textwidth]{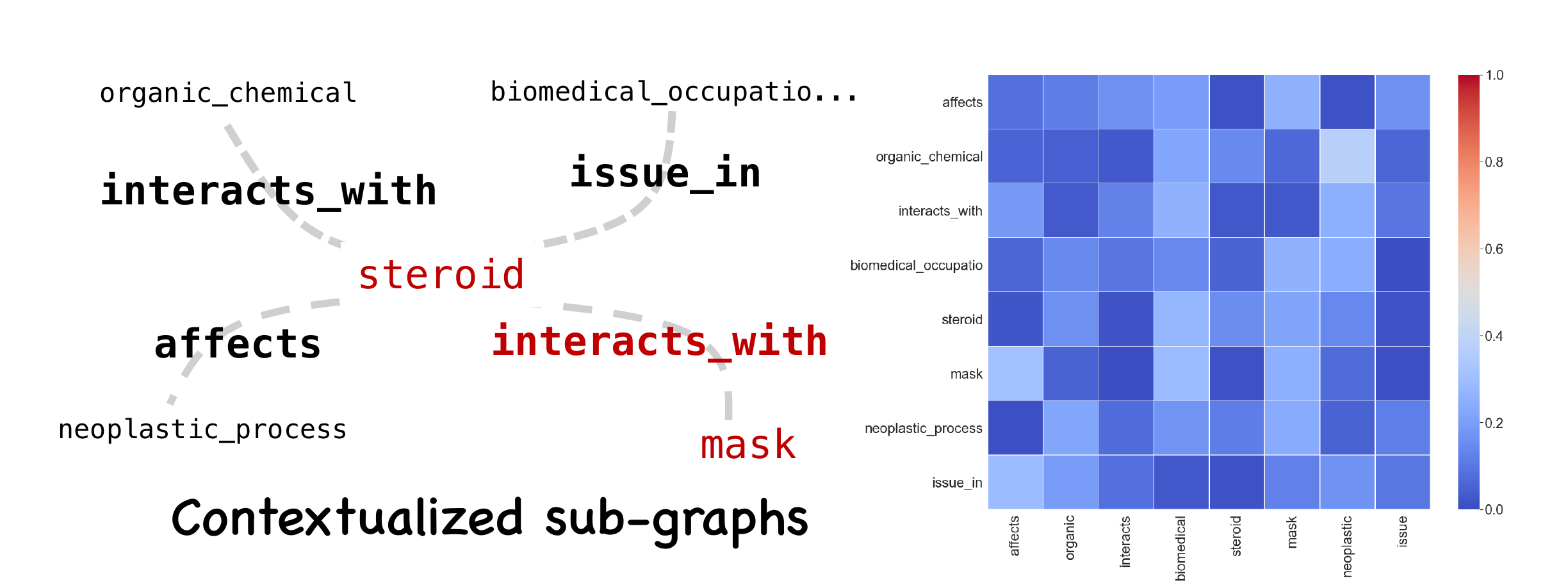}

    \caption{ 
    Attention study on UMLS.
    Left is the contextualized sub-graphs of the center triple (marked in red).
    Right is the attention layer with \emph{structure-enhanced self-attention}.
    }
    \label{fig:matrix}
\end{figure}

\subsubsection{\textbf{Structure-enhanced self-attention}}
To study the effects of structure-enhanced attention in Relphormer, we conduct the ablation study as shown in Figure \ref{fig:others}.
All models without structure-enhanced attention have a performance decay. 
We random an instance and visualize its attention matrix to analyze the impact of structure-enhanced self-attention.
From Figure \ref{fig:matrix}, given a center triple of \textit{(steroid, interacts with, eicosanoid)}, we observe that models with structure-enhanced self-attention have impacts on attentive weights. 
Concretely, injecting structural information with structure-enhanced self-attention can capture the semantic correlation of distance entities.
For example, one entity can learn the structure correlation with other faraway entities in contextualized sub-graph.

\section{Conclusion and Future Work}
In this paper, we propose Relphormer and evaluate its effectiveness in knowledge graph completion, question-answering and recommendation tasks.
As we discussed in Section \ref{sec:Triple2Seq}, Triple2Seq is a simple but effective way to consider context information to address the limitation of input length for the Transformer. 
It is worth studying what contexts play a more critical role in the triple and proposing efficient sampling strategies with an anchor-based approach like NodePiece \cite{DBLP:conf/iclr/0001DWH22} in the future.

\section{Acknowledgments}
We would like to express gratitude to the anonymous reviewers for their kind comments. 
This work was supported by the National Natural Science Foundation of China (No. 62206246), Zhejiang Provincial Natural Science Foundation of China (No. LGG22F030011), Ningbo Natural Science Foundation (2021J190), Yongjiang Talent Introduction Programme (2021A-156-G), CCF-Baidu Open Fund, and Information Technology Center and State Key Lab of CAD\&CG, Zhejiang University.

\bibliographystyle{ACM-Reference-Format}
\bibliography{sample-base}

\end{document}